\documentclass[runningheads]{llncs}
\usepackage[T1]{fontenc}
\usepackage{graphicx}
\usepackage{float}
\usepackage{subcaption}
\usepackage{amsmath}
\usepackage{amssymb}
\usepackage{marvosym}
\usepackage{multirow}
\usepackage{siunitx}
\usepackage{booktabs}
\usepackage[utf8]{inputenc}
\usepackage{tikz,libertine}
\usepackage{xcolor}
\usetikzlibrary{calc,spy}
\tikzstyle{closeup} = [
opacity=1.0,          
height=0.9cm,         
width=0.9cm,          
connect spies, cyan  
]
\tikzstyle{largewindow} = [cyan, line width=0.50mm]
\tikzstyle{smallwindow} = [cyan, line width=0.20mm]
\usepackage{scalerel}
\usepackage{tikz}
\usetikzlibrary{svg.path}

\definecolor{orcidlogocol}{HTML}{A6CE39}
\tikzset{
	orcidlogo/.pic={
		\fill[orcidlogocol] svg{M256,128c0,70.7-57.3,128-128,128C57.3,256,0,198.7,0,128C0,57.3,57.3,0,128,0C198.7,0,256,57.3,256,128z};
		\fill[white] svg{M86.3,186.2H70.9V79.1h15.4v48.4V186.2z}
		svg{M108.9,79.1h41.6c39.6,0,57,28.3,57,53.6c0,27.5-21.5,53.6-56.8,53.6h-41.8V79.1z M124.3,172.4h24.5c34.9,0,42.9-26.5,42.9-39.7c0-21.5-13.7-39.7-43.7-39.7h-23.7V172.4z}
		svg{M88.7,56.8c0,5.5-4.5,10.1-10.1,10.1c-5.6,0-10.1-4.6-10.1-10.1c0-5.6,4.5-10.1,10.1-10.1C84.2,46.7,88.7,51.3,88.7,56.8z};
	}
}

\newcommand\orcidicon[1]{\href{https://orcid.org/#1}{\mbox{\scalerel*{
				\begin{tikzpicture}[yscale=-1,transform shape]
				\pic{orcidlogo};
				\end{tikzpicture}
			}{|}}}}

\newif\ifdraft
\drafttrue

\definecolor{orange}{rgb}{1,0.5,0}
\definecolor{gr}{rgb}{0,0.65,0}

\ifdraft
 \newcommand{\RS}[1]{{\color{red}{\bf RS: #1}}}
 
 \newcommand{\PMN}[1]{{\color{orange}{\bf PMN: #1}}}

\else
 \newcommand{\sout}[1]{}
 \newcommand{\RS}[1]{{\color{red}{}}}
 
 \newcommand{\PMN}[1]{{\color{red}{}}}
 
\fi

\usepackage{hyperref}
\usepackage{color}

\usepackage[nameinlink,capitalise]{cleveref}

\begin{document}
\title{
Geometric Ultrasound Localization Microscopy
}
\titlerunning{G-ULM}
%
\author{Christopher Hahne$^{\textrm{(\Letter)}}$ \and Raphael Sznitman}
%
%
%
\authorrunning{C. Hahne \& R. Sznitman}
%

\institute{ARTORG Center, University of Bern, Switzerland\\ 
\email{christopher.hahne@unibe.ch}}

\maketitle              
\begin{abstract}
%
Contrast-Enhanced Ultra-Sound (CEUS) has become a viable method for non-invasive, dynamic visualization in medical diagnostics, yet Ultrasound Localization Microscopy (ULM) has enabled a revolutionary breakthrough by offering ten times higher resolution. To date, Delay-And-Sum (DAS) beamformers are used to render ULM frames, ultimately determining the image resolution capability. To take full advantage of ULM, this study questions whether beamforming is the most effective processing step for ULM, suggesting an alternative approach that relies solely on Time-Difference-of-Arrival (TDoA) information. To this end, a novel geometric framework for microbubble localization via ellipse intersections is proposed to overcome existing beamforming limitations. We present a benchmark comparison based on a public dataset for which our geometric ULM outperforms existing baseline methods in terms of accuracy and robustness while only utilizing a portion of the available transducer data. %
\keywords{Ultrasound \and Microbubble \and Localization \and Microscopy \and Geometry \and Parallax \and Triangulation \and Trilateration \and Multilateration \and Time-of-Arrival}
\end{abstract}
\section{Introduction}
%
Ultrasound Localization Microscopy (ULM) has revolutionized medical imaging by enabling sub-wavelength resolution from images acquired by piezo-electric transducers and computational beamforming. %
However, the necessity of beamforming for ULM remains questionable. Our work challenges the conventional assumption that beamforming is the ideal processing step for ULM and presents an alternative approach based on geometric reconstruction from Time-of-Arrival (ToA) information. \par

The discovery of ULM has recently surpassed the diffraction-limited spatial resolution and enabled highly detailed visualization of the vascularity~\cite{errico2015ultrafast}. ULM borrows concepts from super-resolution fluorescence microscopy techniques to precisely locate individual particles with sub-pixel accuracy over multiple frames. By the accumulation of all localizations over time, ULM can produce a super-resolved image, providing researchers and clinicians with highly detailed representation of the vascular structure. \par

While Contrast-Enhanced Ultra-Sound (CEUS) is used in the identification of musculoskeletal soft tissue tumours~\cite{marchi:2015}, the far higher resolution capability offered by ULM has great potential for clinical translation to improve the reliability of cancer diagnosis (\textit{i.e.}, enable differentiation of tumour types in kidney cancer~\cite{elbanna:2021} or detect breast cancer tissue~\cite{bar:2021}). %
Moreover, ULM has shown promise in imaging neurovascular activity after visual stimulation (functional ULM)~\cite{renaudin2022functional}.
The pioneering study by Errico~\textit{et al.}~\cite{errico2015ultrafast} initially demonstrated the potential of ULM by successfully localizing contrast agent particles (microbubbles) using a 2D point-spread-function model. In general, the accuracy in MicroBubble (MB) localization is the key to achieving sub-wavelength resolution~\cite{StateOfTheArt}, for which classical imaging methods~\cite{song:2018,heiles2022pala}, as well as deep neural networks~\cite{van2020super,bar:2021}, have recently been reported.
\begin{figure*}[!t]
	\includegraphics[width=\textwidth]{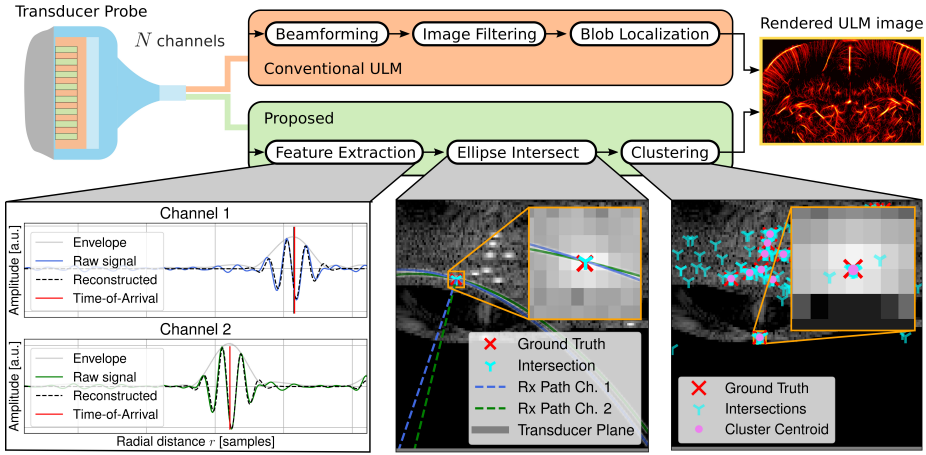}
	\caption{Comparison of ULM processing pipelines: Classical ULM (top) employs computational beamforming from $N$ channels and image filters to localize microbubbles. Our geometric ULM (bottom) consists of a cross-channel phase-consistent Time-of-Arrival detection (left) to form ellipses that intersect at a microbubble position (middle). As a refinement step, ellipse intersections are fused via clustering (right).}\label{fig:pipeline}
\end{figure*}

%
However, the conventional approach for ULM involves using computational beamformers, which may not be ideal for MB localization. For example, a recent study has shown that ultrasound image segmentation can be learned from radio-frequency data and thus without beamforming~\cite{nair2018deep}. Beamforming techniques have been developed to render irregular topologies, whereas MBs exhibit a uniform geometric structure, for which ULM only requires information about its spatial position. %
Although the impact of adaptive beamforming has been studied for ULM to investigate its potential to refine MB localization~\cite{corazza2022beamforming}, optimization of the Point-Spread Function (PSF) poses high demands on the transducer array, data storage, and algorithm complexity.

To this end, we propose an alternative approach for ULM, outlined in Fig.~\ref{fig:pipeline}, that entirely relies on Time-Difference-of-Arrival (TDoA) information, omitting beamforming from the processing pipeline for the first time. We demonstrate a novel geometry framework for MB localization through ellipse intersections to overcome limitations inherent to beamforming. This approach provides a finer distinction between overlapping and clustered spots, improving localization precision, reliability, and computation efficiency.
%
In conclusion, we challenge the conventional wisdom that beamforming is necessary for ULM and propose a novel approach that entirely relies on TDoA information for MB localization. Our proposed approach demonstrates promising results and indicates a considerable trade-off between precision, computation, and memory. 

\section{Method\label{sec:2}}
\begin{figure}[b]
	\centering
	\includegraphics[width=.5\textwidth]{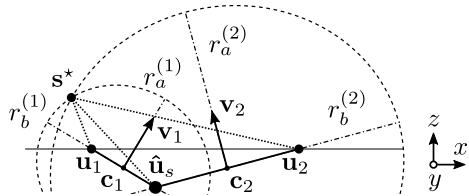}
	\caption{Transducer geometry used for the ellipse intersection and localization of a MB position $\mathbf{s}^{\star}$ from virtual source $\mathbf{\hat{u}}_s$ and receiver positions $\mathbf{u}_n$, which span ellipses rotated by $\mathbf{v}_n$ around their centers $\mathbf{c}_n$.} 
 \label{fig:ellipse_schematic}
\end{figure}

Geometric modeling is a useful approach for locating landmarks in space. One common method involves using a Time-of-Flight (ToF) round-trip setup that includes a transmitter and multiple receivers~\cite{spiel:icra23}. This setup is analogous to the parallax concept in visual imaging, where a triangle is formed between the target, emitter, and receivers, as illustrated in Figure~\ref{fig:ellipse_schematic}. The target's location can be accurately estimated using trilateration by analyzing the time delay between the transmitted and received signals. However, the triangle's side lengths are unknown in the single receiver case, and all possible travel path candidates form triangles with equal circumferences fixed at the axis connecting the receiver and the source. These candidates reside on an elliptical shape. By adding a second receiver, its respective ellipse intersects with the first one resolving the target's 2-D position. Thus, the localization accuracy depends on the ellipse model, which is parameterized by the known transducer positions and the time delays we seek to estimate. 
This section describes a precise echo feature extraction, which is essential for building the subsequent ellipse intersection model. Finally, we demonstrate our localization refinement through clustering. \par
\subsection{Feature Extraction}
Feature extraction of acoustic signals has been thoroughly researched~\cite{Zonzini:2022,Hahne:22}. To leverage the geometric ULM localization, we wish to extract Time-of-Arrival (ToA) information (instead of beamforming) at sub-wavelength precision. Despite the popularity of deep neural networks, which have been studied for ToA detection~\cite{Zonzini:2022}, we employ an energy-based model~\cite{Hahne:22} for echo feature extraction to demonstrate the feasibility of our geometric ULM at the initial stage. Ultimately, future studies can combine our proposed localization with a supervised network. Here, echoes $f(\mathbf{m}_k; t)$ are modeled as Multimodal Exponentially-Modified Gaussian Oscillators~(MEMGO)~\cite{Hahne:22}, %

\begin{align}
f(\mathbf{m}_k;t)= \alpha_k\,\exp\left(-\frac{\left(t-\mu_k\right)^2}{2\sigma_k^2}\right) \left(1+ \text{erf}\left(\eta_k\frac{t-\mu_k}{\sigma_k\sqrt{2}}\right)\right) \cos\left(\omega_k \left(t - \mu_k\right) + \phi_k\right), 
\label{eq:model}
\end{align}
where $t \in \mathbb{R}^{T}$ denotes the time domain with a total number of $T$ samples and \linebreak \mbox{$\mathbf{m}_k=\left[\alpha_k,\mu_k,\sigma_k,\eta_k,\omega_k, \phi_k\right]^\intercal\in\mathbb{R}^{6}$} contains the amplitude~$\alpha_k$, mean~$\mu_k$, spread~$\sigma_k$, skew~$\eta_k$, angular frequency~$\omega_k$ and phase~$\phi_k$ for each echo $k$. Note that $\text{erf}(\cdot)$ is the error function.
To estimate these parameters iteratively, the cost function is given by,
\begin{align}
\mathcal{L}_{\text{E}}\left(\mathbf{\hat{m}}_n\right)=\left\lVert y_n(t)-\sum_{k=1}^K f\left(\mathbf{m}_k;t\right)\right\rVert_2^2, 
\label{eq:minimization}
\end{align}
where $y_n(t)$ is the measured signal from waveform channel $n\in\{1, 2, \dots, N\}$ and the sum over $k$ accumulates all echo components $\mathbf{\hat{m}}_n=[\mathbf{m}_1^{\intercal}, \mathbf{m}_2^{\intercal}, \dots, \mathbf{m}_K^{\intercal}]^{\intercal}$. We get the best echo feature set $\mathbf{\hat{m}}_n^\star$ over all iterations $j$ via,
\begin{align}
\mathbf{\hat{m}}_n^\star=\underset{\mathbf{\hat{m}}_n^{(j)}}{\operatorname{arg\,min}} \, \left\{\mathcal{L}_{\text{E}}\left(\mathbf{\hat{m}}_n^{(j)}\right)\right\},
\label{eq:goal}
\end{align}
for which we use the Levenberg-Marquardt solver. Model-based optimization requires initial estimates to be nearby the solution space. For this, we detect initial ToAs via gradient-based analysis of the Hilbert-transformed signal to set $\mathbf{\hat{m}}_n^{(1)}$ as in~\cite{Hahne:22}. \par

Before geometric localization, one must ensure that detected echo components correspond to the same MB. In this work, echo matching is accomplished in a heuristic brute-force fashion. Given an echo component $\mathbf{m}^\star_{n,k}$ from a reference channel index $n$, a matching echo component from an adjacent channel index $n\pm g$ with gap $g\in\mathbb{N}$ is found by $k+h$ in the neighborhood of $h\in\{-1, 0, 1\}$. %
A corresponding phase-precise ToA $t^\star_{n,k}$ is obtained by 
$
    t^\star_{n\pm g,k}=\mu^\star_{n\pm g,k+h}+\phi^\star_{n,k}-\Delta
$
, which takes $\mu^\star_{n,k}$ and $\phi^\star_{n,k}$ from $\mathbf{\hat{m}}_n^\star$ for phase-precise alignment across transducer channels after upsampling. Here, $\Delta$ is a fixed offset to accurately capture the onset of the MB locations~\cite{Jeffries:2017}. We validate echo correspondence through a re-projection error in adjacent channels and reject those with weak alignment. %
\subsection{Ellipse Intersection}
While ellipse intersections can be approximated iteratively, 
we employ Eberly's closed-form solution~\cite{EberlyEllipse} owing to its fast computation property. Although one might expect that the intersection of arbitrarily placed ellipses is straightforward, it involves advanced mathematical modelling due to the degrees of freedom in the ellipse positioning. 
%
%
An ellipse is drawn by radii $(r_a,r_b)$ of the major and minor axes with,
\begin{align}
r_a = \frac{t^\star_{n,k}}{2},
\quad \text{and} \quad
r_b = \frac{1}{2} \, \sqrt{\left(t^\star_{n,k}\right)^2-\left\lVert\mathbf{\hat{u}}_s-\mathbf{u}_n\right\rVert^2_2},
\end{align}
where the virtual transmitter $\mathbf{\hat{u}}_s \in \mathbb{R}^2$ and each receiver $\mathbf{u}_n \in \mathbb{R}^2$ with channel index $n$ represent the focal points of an ellipse, respectively. %
For the intersection, we begin with the ellipse standard equation. Let any point $\mathbf{s}\in\mathbb{R}^2$ located on an ellipse and displaced by its center $\mathbf{c}_n\in\mathbb{R}^2$ such that,
\begin{equation}
\left(\mathbf{s}-\mathbf{c}_n\right)^{\intercal}
\left(\frac{\lVert\mathbf{v}_n\rVert_2^2}{r_a^2|\mathbf{v}_n|^2}
+
\frac{\lVert\mathbf{v}_n^{\perp}\rVert_2^2}{r_b^2|\mathbf{v}_n^{\perp}|^2}\right)
\left(\mathbf{s}-\mathbf{c}_n\right) \\
=
\left(\mathbf{s}-\mathbf{c}_n\right)^{\intercal}
\mathbf{M}
\left(\mathbf{s}-\mathbf{c}_n\right)
= 1,
\label{eq:standard}
\end{equation}
where $\mathbf{M}$ contains the ellipse equation with $\mathbf{v}_n$ and $\mathbf{v}^{\perp}_n$ as a pair of orthogonal ellipse direction vectors, corresponding to their radial extents $(r_0,r_1)$ as well as the squared norm $\lVert\cdot\rVert^2_2$ and vector norm $|\cdot|$.
%
For subsequent root-finding, it is the goal to convert the standard Eq.~\eqref{eq:standard} to a quadratic polynomial with coefficients $b_j$ given by,
$B(x,y)=b_0+b_1x+b_2y+b_3x^2+b_4xy=0$, which, when written in vector-matrix form reads,
\begin{equation}
0
=
\begin{bmatrix}x & y\end{bmatrix}
\begin{bmatrix}
b_3 & b_4/2 \\ b_4/2 & b_5
\end{bmatrix}
\begin{bmatrix} x \\ y\end{bmatrix}
+
\begin{bmatrix} b_1 & b_2\end{bmatrix}
\begin{bmatrix} x \\ y\end{bmatrix}
+ b_0 =
\mathbf{s}^{\intercal}\mathbf{B}\mathbf{s}+\mathbf{b}^{\intercal}\mathbf{s}+b_0,
\label{eq:intersect_algebra}
\end{equation}
where $\mathbf{B}$ and $\mathbf{b}$ carry high-order polynomial coefficients $b_j$ found via matrix factorization~\cite{EberlyEllipse}. An elaborated version of this is found in the supplementary material. \par
%
%
Let two intersecting ellipses be given as quadratic equations $A(x,y)$ and $B(x,y)$ with coefficients $a_j$ and $b_j$, respectively. Their intersection is found via polynomial root-finding of the equation,
\begin{equation}
D(x,y)=d_0+d_1x+d_2y+d_3x^2+d_4xy=0,
\label{eq:dk}
\end{equation}
where $\forall j, \, d_j=a_j-b_j$. When defining $y=w-(a_2+a_4x)/2$ to substitute $y$, we get
$
A(x,w)=w^2+(a_0+a_1x+a_3x^2)-(a_2+a_4x)^2/4=0
$
which after rearranging is plugged into~\eqref{eq:dk} to yield an
intersection point $\mathbf{s}_i^\star=[x_i,w_i]^\intercal$. We refer the interested reader to the insightful descriptions in~\cite{EberlyEllipse} for further implementation details. 
\subsection{Clustering}
Micro bubble reflections are dispersed across multiple waveform channels yielding groups of location candidates for the same target bubble. Localization deviations result from ToA variations, which can occur due to atmospheric conditions, receiver clock errors, and system noise. Due to the random distribution of corresponding ToA errors~\cite{errico2015ultrafast}, we regard these candidates as clusters. Thus, we aim to find a centroid $\mathbf{p}^\star$ of each cluster using multiple bi-variate probability density functions of varying sample sizes by, %
\begin{align}
m(\mathbf{p}^{(j)}) = \frac{ \sum_{\mathbf{s}^{\star}_i \in \mathbf{\Omega}^{(j)}} \exp\left(\lVert\mathbf{s}^{\star}_i - \mathbf{p}^{(j)}\rVert^2_2\right) \mathbf{s}^{\star}_i } {\sum_{\mathbf{s}^{\star}_i \in \mathbf{\Omega}^{(j)}} \exp\left(\lVert\mathbf{s}^{\star}_i - \mathbf{p}^{(j)}\rVert^2_2\right)}
\end{align}
Here, the bandwidth of the kernel is set to $\lambda/4$. The Mean Shift algorithm updates the estimate $\mathbf{p}^{(j)}$ by setting it to the weighted mean density on each iteration $j$ until convergence. In this way, we obtain the position of the target bubble.
\section{Experiments\label{sec:3}}
%

\subsubsection{Dataset:}
We demonstrate the feasibility of our geometric ULM and present benchmark comparison outcomes based on the PALA dataset~\cite{heiles2022pala}. This dataset is chosen as it is publicly available, allowing easy access and reproducibility of our results. To date, it is the only public ULM dataset featuring Radio Frequency (RF) data as required by our method. 
Its third-party simulation data makes it possible to perform a numerical quantification and direct comparison of different baseline benchmarks for the first time, which is necessary to validate the effectiveness of our proposed approach. 

\subsubsection{Metrics:}
For MB localization assessment, the minimum Root Mean Squared Error (RMSE) between the estimated $\mathbf{p}^{\star}$ and the nearest ground truth position is computed. To align with the PALA study~\cite{heiles2022pala}, only RMSEs less than $\lambda/4$ are considered true positives and contribute to the total RMSE of all frames. In cases where the RMSE distance is greater than $\lambda/4$, the estimated $\mathbf{p}^{\star}$ is a false positive. Consequently, ground truth locations without an estimate within the $\lambda/4$ neighbourhood are false negatives. We use the Jaccard Index to measure the MB detection capability, which considers both true positives and false negatives and provides a robust measure of each algorithm's performance. The Structural Similarity Index Measure (SSIM) is used for image assessment. %

For a realistic analysis, we employ the noise model used in~\cite{heiles2022pala}, which is given by,
\begin{align}
n(t)\sim\mathcal{N}(0, \sigma_{p}^2) \times \max(y_n(t))\times10^{(L_{A}+L_{C})/20}\pm\max(y_n(t))\times10^{L_{C}/20}, \label{eq:noise}
\end{align}
where $\sigma_{p}=\sqrt{B\times10^{P/10}}$ and $\mathcal{N}(0, \sigma_{p}^2)$ are normal distributions with mean 0 and variance $\sigma_{p}^2$. Here, $L_{C}$ and $L_{A}$ are noise levels in dB, and $n(t)$ is the array of length $T$ containing the random values drawn from this distribution. The additive noise model is then used to simulate a waveform channel $y'_n(t)=y_n(t)+n(t)\circledast g(t,\sigma_f)$ suffering from noise, where $\circledast$ represents the convolution operator, and $g(t,\sigma_f)$ is the one-dimensional Gaussian kernel with standard deviation $\sigma_f=1.5$. 
To mimic the noise reduction achieved through the use of sub-aperture beamforming with 16 transducer channels~\cite{heiles2022pala}, we multiplied the RF data noise by a factor of 4 for an equitable comparison.

\subsubsection{Baselines:}
We compare our approach against state-of-the-art methods that utilize beamforming together with classical image filterings~\cite{errico2015ultrafast}, Spline interpolation~\cite{song:2018}, Radial Symmetry~(RS)~\cite{heiles2022pala} and a deep-learning-based U-Net~\cite{van2020super} for MB localization. To only focus on the localization performance of each algorithm, we conduct the experimental analysis without temporal tracking. We obtain the results for classical image processing approaches directly from the open-source code provided by the authors of the PALA dataset~\cite{heiles2022pala}. As there is no publicly available implementation of~\cite{van2020super} to date, we model and train the U-Net~\cite{ronneberger2015u} according to the paper description, including loss design, layer architecture, and the incorporation of dropout. Since the U-Net-based localization is a supervised learning approach, we split the PALA dataset into sequences 1-15 for testing and 16-20 for training and validation, with a split ratio of 0.9, providing a sufficient number of 4500 training frames. 
\subsubsection{Results:}
Table~\ref{tab:benchmark} provides the benchmark comparison results with state-of-the-art methods. Our proposed geometric inference indicates the best localization performance represented by an average RMSE of around one-tenth of a wavelength. Also, the Jaccard Index reflects an outperforming balance of true positive and false negative MB detections by our approach. These results support the hypothesis that our proposed geometric localization inference is a considerable alternative to existing \mbox{beamforming-based} methods. %
Upon closer examination of the channels column in Table~\ref{tab:benchmark}, it becomes apparent that our geometric ULM achieves reasonable localization performance with only a fraction of the 128 channels available in the transducer probe. Using more than 32 channels improves the Jaccard Index but at the expense of computational resources. This finding confirms the assumption that transducers are redundant for MB tracking. The slight discrepancy in SSIM scores between our 128-channel results and the 64-channel example may be attributed to the higher number of false positives in the former, which decreases the overall SSIM value. 
\begin{table}[t]
	\centering
	\caption{Summary of localization results using $15$k frames of the PALA dataset~\cite{heiles2022pala}. The RMSE is reported as mean±std, best scores are bold and units are given in brackets.}\label{tab:benchmark}
	\begin{tabular}{lccccc}
		\toprule
		Method & Channels [\#] & RMSE [$\lambda/10$] & Jaccard [$\%$] & Time [s] & SSIM [$\%$] \\
		\midrule
        Weighted Avg.~\cite{heiles2022pala} & 128 & $1.287\pm0.162$ & 44.253 & 0.080 & 69.49 \\
        Lanczos~\cite{heiles2022pala} & 128 & $1.524\pm0.175$ & 38.688 & 0.382 & 75.87 \\
		RS~\cite{heiles2022pala} & 128 & $1.179 \pm 0.172$ & 50.330 & 0.099 & 72.17 \\ 
        Spline~\cite{song:2018} & 128 & $1.504\pm0.174$ & 39.370 & 0.277 & 75.72 \\
		2-D Gauss Fit~\cite{song:2018} & 128 & $1.240 \pm 0.162$
		 & 51.342 & 3.782 & 73.93 \\
		U-Net~\cite{van2020super} & 128 &  $1.561 \pm 0.154$ & 52.078 & \bf{0.004} & 90.07 \\
		\hline
		\multirow{5}{*}{\shortstack{G-ULM \\ (proposed)}}
		 & 8 & $1.116\pm0.206$ & 38.113 &  0.268 & 79.74 \\	
		 & 16 & $1.077\pm0.136$ & 66.414 & 0.485 & 87.10 \\
		 & 32 & $1.042 \pm 0.125$ & 72.956 & 0.945 & 92.18 \\
		 & 64 & $1.036 \pm 0.124$ & 73.175 & 1.317 & \textbf{93.70} \\
		 & 128 & $\bf{0.967\pm0.109}$ & \textbf{78.618} & 3.747 & 92.02 \\
		\bottomrule
	\end{tabular}
\end{table}

\begin{figure}[!ht]
  \centering
  \begin{minipage}[b]{0.242\textwidth}
    \centering
    \includegraphics[width=\textwidth,trim=240 538 1030 238, clip]{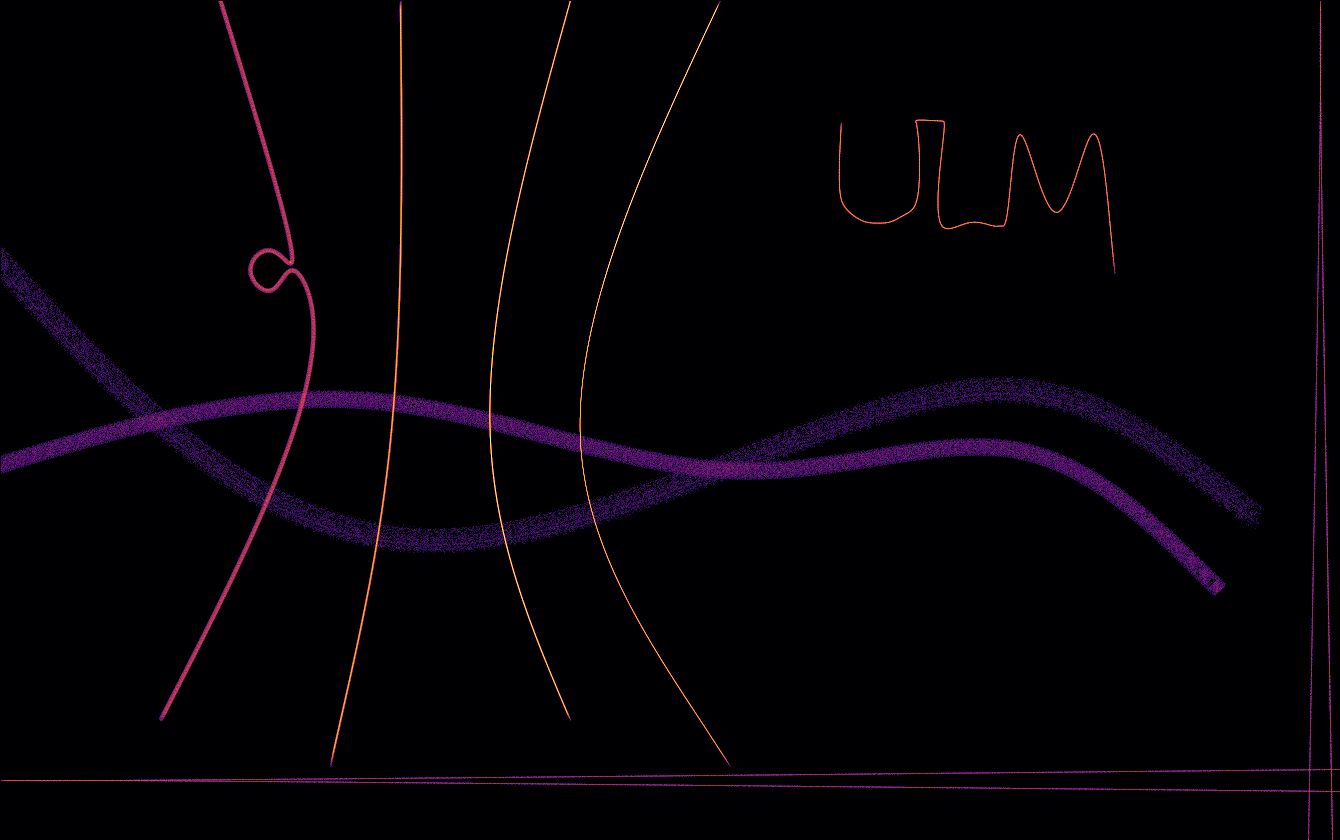}\par\vspace{1mm}
    \includegraphics[width=\textwidth,trim=310 45 980 750, clip]{./img/gtru_ulm_img_15k_gam0.9.png}\par\vspace{1mm}
    \includegraphics[width=\textwidth,trim=820 560 200 100, clip]{./img/gtru_ulm_img_15k_gam0.9.png}
    \subcaption{Ground Truth\label{fig:method1}}
    \vspace{+.1cm}
  \end{minipage}
  \hfill
  \begin{minipage}[b]{0.242\textwidth}
    \centering
    \includegraphics[width=\textwidth,trim=240 538 1030 238, clip]{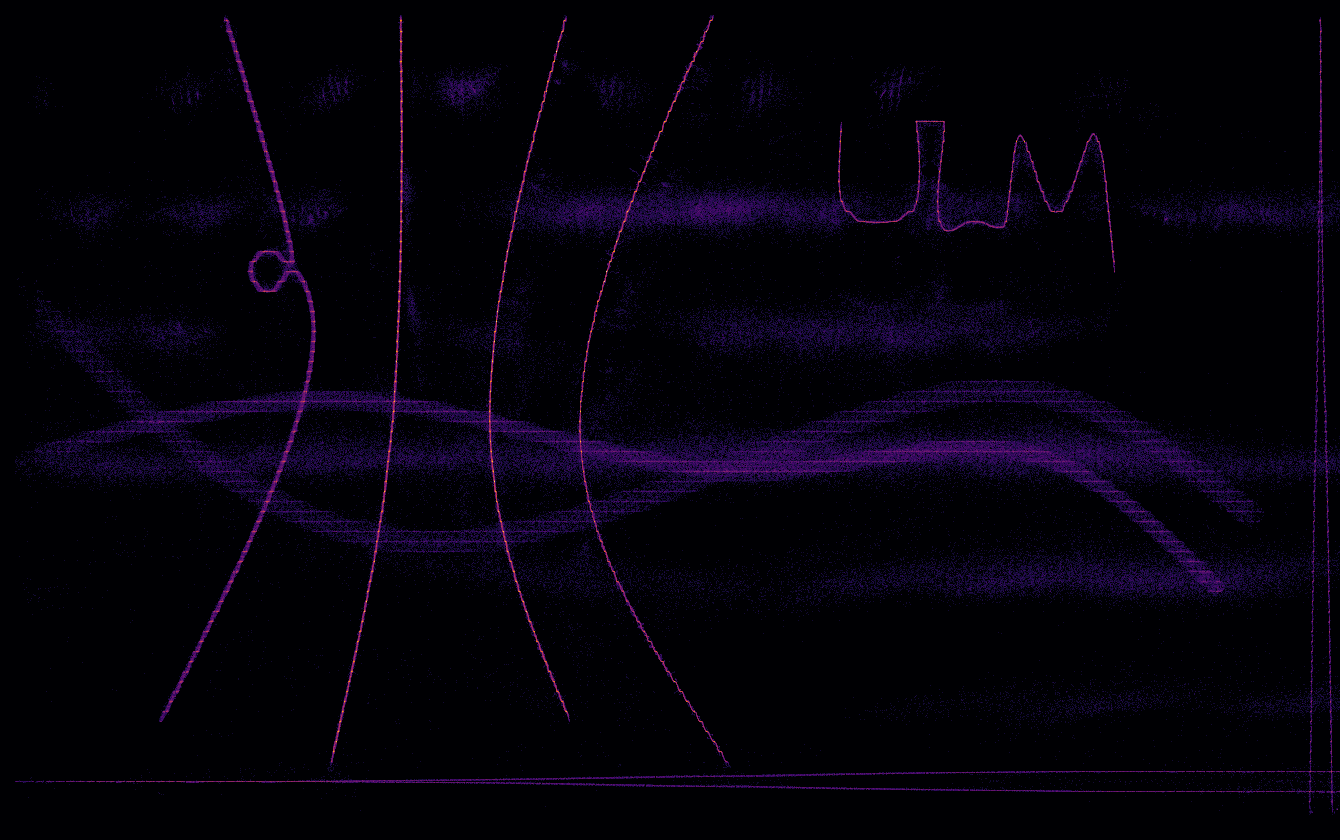}\par\vspace{1mm}
    \includegraphics[width=\textwidth,trim=310 45 980 750, clip]{./img/pala_ulm_img_15k_gam0.9_128chs.png}\par\vspace{1mm}
    \includegraphics[width=\textwidth,trim=820 560 200 100, clip]{./img/pala_ulm_img_15k_gam0.9_128chs.png}
    \subcaption{RS~\{128\}~\cite{heiles2022pala}\label{fig:method2}}
    \vspace{+.1cm}
  \end{minipage}
  \hfill
  \begin{minipage}[b]{0.242\textwidth}
    \centering
    \includegraphics[width=\textwidth,trim=240 538 1030 238, clip]{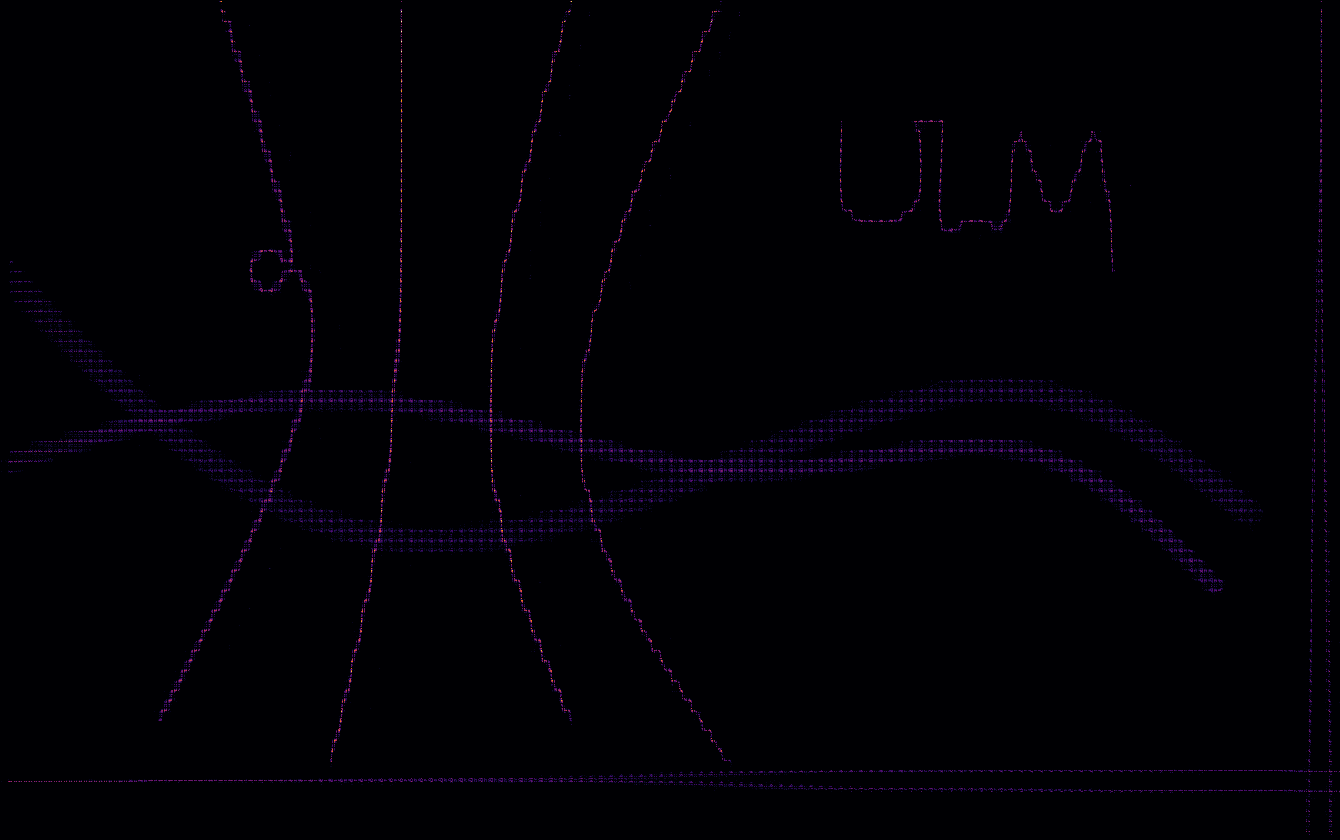}\par\vspace{1mm}
    \includegraphics[width=\textwidth,trim=310 45 980 750, clip]{./img/unet_ulm_img_15k_128chs_gam0.9.png}\par\vspace{1mm}
    \includegraphics[width=\textwidth,trim=820 560 200 100, clip]{./img/unet_ulm_img_15k_128chs_gam0.9.png}
    \subcaption{U-Net~\{128\}~\cite{van2020super}\label{fig:method3}}
    \vspace{+.1cm}
  \end{minipage}
  \hfill
  \begin{minipage}[b]{0.242\textwidth}
    \centering
    \includegraphics[width=\textwidth,trim=240 538 1030 238, clip]{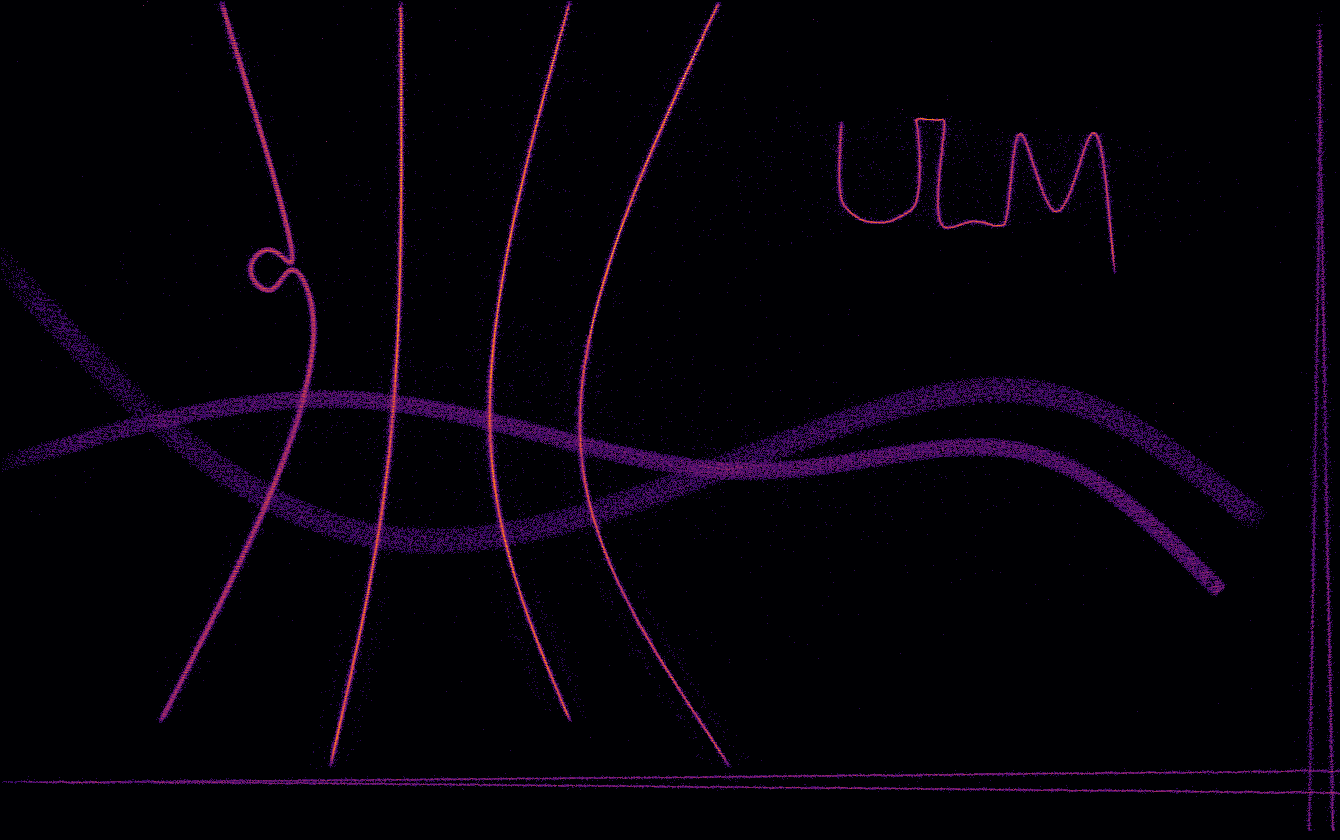}\par\vspace{1mm}
    \includegraphics[width=\textwidth,trim=310 45 980 750, clip]{./img/pace_ulm_img_15k_gam0.9_16chs.png}\par\vspace{1mm}
    \includegraphics[width=\textwidth,trim=820 560 200 100, clip]{./img/pace_ulm_img_15k_gam0.9_16chs.png}
    \subcaption{Ours~\{16\}\label{fig:method4}}
    \vspace{+.1cm}
  \end{minipage}
  \begin{minipage}[b]{\textwidth}
    \includegraphics[width=\textwidth,trim=7cm 20.5cm 3cm 2.5cm,clip]{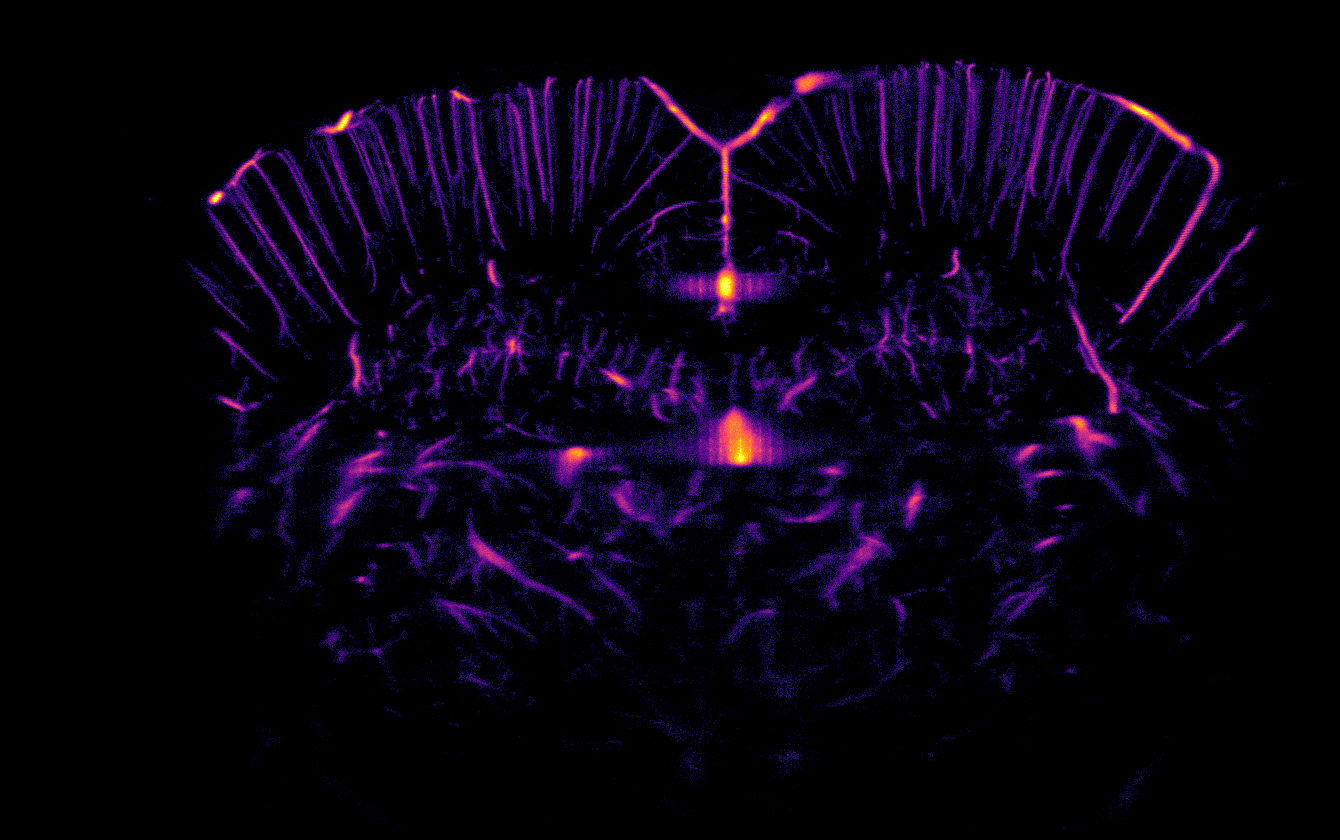}
    \subcaption{Rat brain in vivo result by our proposed method \{64\}\label{fig:invivo}}
  \end{minipage}
    \caption{Rendered ULM regions from Table~\ref{tab:benchmark} in (\subref{fig:method1}) to (\subref{fig:method4}) and a rat brain result in (\subref{fig:invivo}) without temporal tracking. Numbers in curly brackets indicate the transducer number.} 
    \label{fig:render_ulm_crop}
\end{figure}
We provide rendered ULM image regions for visual inspection in Fig.~\ref{fig:render_ulm_crop} with full frames in the supplementary material. To enhance visibility, all images are processed with sRGB and additional gamma correction using an exponent of 0.9. The presence of noisy points in \cref{fig:method2,fig:method3,fig:method4} is attributed to the excessive false positive localizations, resulting in poorer SSIM scores. Overall, these visual observations align with the numerical results presented in Table~\ref{tab:benchmark}.
%
An NVIDIA RTX2080 GPU was used for all computations and time measurements. To improve performance, signal processing chains are often pipelined, allowing for the simultaneous computation of subsequent processes. Table~\ref{tab:benchmark} lists the most time-consuming process for each method, which acts as the bottleneck. For our approach, the MEMGO feature extraction is the computationally most expensive process, followed by clustering. However, our method contributes to an overall efficient computation and acquisition time, as it skips beamforming and coherent compounding~\cite{montaldo2009coherent} with the latter reducing the capture interval by two-thirds. 

Table~\ref{tab:noise_benchmark} presents the results for the best-of-3 algorithms at various noise levels $L_C$. As the amount of noise from \eqref{eq:noise} increases, there is a decline in the Jaccard Index, which suggests that each method is more susceptible to false detections from noise clutter. Although our method is exposed to higher noise in the RF domain, it is seen that $L_C$ has a comparable impact on our method. However, it is important to note that the U-Net yields the most steady and consistent results for different noise levels. 

\begin{table}[h!]
	\centering
	\caption{
    Performance under noise variations from 128 (others) vs  16 (ours) transducers.
    }\label{tab:noise_benchmark}
	\begin{tabular}{c c c c c c c}
		\toprule
		Noise & \multicolumn{3}{c}{RMSE [$\lambda/10$]} & \multicolumn{3}{c}{Jaccard Index [$\%$]} 
        \\
		\midrule
        $L_{C}$ [dB] & RS~\cite{heiles2022pala} & U-Net~\cite{van2020super} & Ours & RS~\cite{heiles2022pala} & U-Net~\cite{van2020super} & Ours 
        \\
        \hline
		-30 & $1.245\pm0.171$ & $1.564\pm0.151$ & $1.076\pm0.136$ & 54.036 & 51.032 & 65.811 \\
  		-20 & $1.496\pm0.223$ & $1.459\pm0.165$ & $1.262\pm0.249$ & 27.037 & 45.647 & 27.962 
        \\
        -10 & $1.634\pm0.542$ & $1.517\pm0.238$ & $1.459\pm0.564$ & 2.510 & \text{18.162} & 3.045 \\
	    \bottomrule
 \end{tabular}
\end{table}
\section{Summary\label{sec:4}}
This study explored whether a geometric reconstruction may serve as an alternative to beamforming in ULM. %
We employed an energy-based model for feature extraction in conjunction with ellipse intersections and clustering to pinpoint contrast agent positions from RF data available in the PALA dataset. %
We carried out a benchmark comparison with state-of-the-art methods, demonstrating that our geometric model provides enhanced resolution and detection reliability with fewer transducers. %
This capability will be a stepping stone for 3-D ULM reconstruction where matrix transducer probes typically consist of 32 transducers per row only. %
It is essential to conduct follow-up studies to evaluate the high potential of our approach in an extensive manner before entering a pre-clinical phase. The promising results from this study motivate us to expand our research to more RF data scenarios. We believe our findings will inspire further research in this exciting and rapidly evolving field. %

\subsubsection{Acknowledgments:} This research is supported by the Hasler Foundation under project number 22027.


%
%
\bibliography{paper1191}
\bibliographystyle{splncs04}
\end{document}